\documentclass{article}
\usepackage{ijcai20}

\usepackage{times}
\usepackage{soul}
\usepackage{url}
\usepackage[hidelinks]{hyperref}
\usepackage[utf8]{inputenc}
\usepackage[small]{caption}
\usepackage{graphicx}
\usepackage{amsmath}
\usepackage{amsthm}
\usepackage{booktabs}
\usepackage{algorithm}
\usepackage[noend]{algorithmic}
\usepackage{eqparbox}

\usepackage{lipsum}
\usepackage{cleveref}
\graphicspath{{images/}}

\title{Injecting Domain Knowledge in Neural Networks:\\ a Controlled Experiment on a Constrained Problem}

\author{
Mattia Silvestri\and
Michele Lombardi\And
Michela Milano
\affiliations
DISI, University of Bologna
\emails
\{mattia.silvestri4,michele.lombardi2,michela.milano\}@unibo.it
}

\begin{document}

\maketitle

\begin{abstract}
Given enough data, Deep Neural Networks (DNNs) are capable of learning complex input-output relations with high accuracy. In several domains, however, data is scarce or expensive to retrieve, while a substantial amount of expert knowledge is available. It seems reasonable that if we can inject this additional information in the DNN, we could ease the learning process.
One such case is that of Constraint Problems, for which declarative approaches exists and pure ML solutions have obtained mixed success. Using a classical constrained problem as a case study, we perform controlled experiments to probe the impact of progressively adding domain and empirical knowledge in the DNN. Our results are very encouraging, showing that (at least in our setup) embedding domain knowledge at training time can have a considerable effect and that a small amount of empirical knowledge is sufficient to obtain practically useful results.
\end{abstract}

\section{Introduction}%
\label{sec:Introduction}

Given enough data, Deep Neural Networks (DNNs) are capable of learning complex input-output relations with high accuracy. In many domains, however, there exists also a substantial degree of expert knowledge: it seems reasonable that if we can inject this additional information in the DNN, we could ease the learning process. Indeed, methods for hybridizing learning and reasoning (or for taking into account constraints at training time) can accelerate convergence or improve the accuracy, especially when supervised data is scarce.

In this paper we aim at characterizing this trade-off between implicit knowledge (derived from data) and explicit knowledge (supplied by experts), via a set of controlled experiments. On this purpose, we use a setting that is both rigorous enough from a scientific standpoint and practically relevant: that of constrained problems.

Constrained problems involve assigning values to a set of variables, subject to a number of constraints, and possibly with the goal of minimizing a cost metric. Depending on the lack or presence of a cost function, they are formally known as Constraint Satisfaction Problems (CSPs) or Constraint Optimization problems (COPs).

Constrained problem are classically modeled by domain experts in a fully declarative fashion: however, such models can be hard to design, may rely on simplistic and unquantifiable approximations, and may fail to take into account constraints (or preferences) that are not known to the expert, despite being satisfied in historical solutions. Data-driven methods for constrained problems offer a potential solution for some of these issues, but they may have trouble maintaining feasibility and they struggle with the (very) limited number of past solutions available for practical use cases.


We use as a benchmark the Partial Latin Square (PLS) completion problem, which requires to complete a partially filled $n \times n$ square with values in $\{1..n\}$, such that no value appears twice on any row or column. Despite its simplicity, the PLS is NP-hard, unless we start from an empty square; the problem has practical applications (e.g. in optical fiber routing), and serves as the basis for more complex problems (e.g. timetabling). Using a classical constrained problem as a case study grants access to reliable domain knowledge (the declarative formulation), and facilitates the generation of empirical data (problem solutions). This combination enables controlled experiments that are difficult to perform on more traditional datasets.

We train a problem-agnostic, data-driven, solution approach on a pool of solutions, and we inject domain knowledge (constraints) both at training time and at solution generation time. We then adjust the amount of initial data (empirical knowledge) and of injected constraints (domain knowledge), and assess the ability of the approach to yield feasible solutions. Our results are very encouraging, showing that (at least in our setup) \emph{embedding domain knowledge in a data-driven approach can have a considerable effect, and that a small amount of empirical knowledge is sufficient to obtain practically useful results}.

As a byproduct of our analysis, we develop \emph{general techniques for taking into account constraints in data-driven methods for decision problems, based on easily accessible methods} from the Constraint Programming and Machine Learning domains. While such techniques are originally designed for problems with discrete decision, they should be adaptable to numeric decisions as well. Hence, despite our focus remains on a scientific investigation, we also regard this paper as a relevant step towards practical applicability for some data-driven solution methods for constrained problems.

The paper is organized as follows: \Cref{sec:Related Works} briefly surveys the related literature and motivates the choice of our baseline techniques; \Cref{sec:Methods} discusses the details of the problem and methods we use; \Cref{sec:Empirical Analysis} presents the results of our analysis, while \Cref{sec:Conclusion} provides concluding remarks.

\section{Related Works and Baseline Choice}%
\label{sec:Related Works}

The analysis that we aim to perform requires \emph{1) a data-driven technique that can solve a constrained problem, with no access to its structure}; moreover, we need \emph{2) methodologies for injecting domain knowledge in such a system, both at training time and after training}. In this section, we briefly survey methods available in the literature for such tasks and we motivate our selection of techniques.

\paragraph{Neural Networks for Solving Constrained Problems}

The integration of Machine Learning methods for the solution of constrained problems is an active research topic, which has recently been surveyed in \cite{bengio2018machine}. Many such approaches consider how ML can improve specific steps of the solution process: here, however, we are interested in methods that use learning to replace (entirely or in part) the modeling activity itself. These include Constraint Acquisition approaches (e.g. \cite{DBLP:journals/ai/BessiereKLO17}), which attempt to learn a declarative problem description from feasible/infeasible variable assignments. These approaches may however have trouble dealing with implicit knowledge (e.g. preferences) that cannot be easily stated in a well defined constraint language. Techniques for encoding Machine Learning models in constrained problems (e.g. \cite{FISCHETTICPAIOR2018,DBLP:journals/ai/LombardiMB17,VERWER2017368,mivsic2017optimization}) are capable of integrating empirical and domain knowledge, but not at training time; additionally, they require to know a-priori which variables are involved in the constraints to be learned.

Some approaches (e.g. \cite{5726908,7804991}) rely on carefully structured Hopfield Networks to solve constrained problems, but designing these networks (or their training algorithms) requires full problem knowledge. Recently, reinforcement learning and Pointer Networks \cite{bello2016neural} or Attention \cite{kool2018attention} have been used for solving specific classes of constrained problems, with some measure of success. These approaches also require a high degree of problem knowledge to generate the reward signal, and to some degree for the network design. The method from \cite{xu2018towards} applies Neural Networks to predict the feasibility of a binary CSP, with a very high degree of accuracy; the prediction is however based on a representation of the allowed variable-value pairs, and hence requires explicit information about the problem.

\emph{To the best of the authors knowledge, the only example of problem-agnostic, data-driven, approach for the solution of constrained problems is the one in \cite{galassi2018-01}}. Here, a Neural Network is used to learn how to extend a partial variable assignment so as to retain feasibility. Despite its limited practical effectiveness, such method shares the best properties of Constraint Acquisition (no explicit problem information), without being restricted to constraints expressed in a classical declarative language. We therefore chose this approach as our baseline.

\paragraph{Domain Knowledge in Neural Networks}

There are several approaches for incorporating external knowledge in Neural Networks, none of which has been applied so far on constrained decision problems. One method to do take into account domain knowledge \emph{at training time} is the so-called Semantic Based Regularization \cite{diligenti2017semantic}, which is based on the idea of converting constraints into regularizing terms in the loss function used by a gradient-descent algorithm. Other techniques include Logic Tensor Networks (LTNs) \cite{serafini2016logic}, which replace the entire loss function with a fuzzy formula defined on logical predicates. LTNs are connected to Differentiable Reasoning \cite{vanKrieken2019semi}, which uses relational background knowledge to benefit from unlabeled data. Domain knowledge has also been introduced in differentiable Machine Learning (in particular Deep Networks) by properly designing their structure, rather than the loss function: examples include Deep Structured Models (see e.g. \cite{lin2016efficient} and \cite{Xuezhe2016}, the latter integrating deep learning with Conditional Random Fields).

Integration of external knowledge in Neural Networks \emph{after training} is considered for example in DeepProbLog~\cite{manhaeve2018deepproblog}, where DNNs with probabilistic output (classifiers in particular) are treated as predicates. Markov Logic Networks achieve a similar results via the use of Markov Fields defined over First Order Logic formulas \cite{richardson2006markov}, which may be defined via probabilistic ML models. \cite{rocktaschel2017end} presents a Neural Theorem Prover using differentiable predicates and the Prolog backward chaining algorithm.

Recent works such as \cite{marra2019integrating} are capable of integrating probabilistic reasoning and Neural Networks both during and after training. Even more general is the Differentiable Inductive Logic approach \cite{evans2018learning}, which proposes a framework that can solve ML tasks typical of traditional Inductive Logic Programming systems, but is also robust to noise and error in the training data.

\emph{We use a method loosely based on Semantic Based Regularization for injecting knowledge} at training time, as it offers a good compromise between flexibility and simplicity. For injecting knowledge \emph{after training, we rely on a very simple method well suited for constrained problems} (i.e. using constraint propagation to adjust the output of the ML model).

\section{Basic Methods}
\label{sec:Methods}

We reimplement the approach from \cite{galassi2018-01} and extend it via number of techniques, described in this section together with our evaluation procedure. The actual setup and results of our experiments are presented in \Cref{sec:Empirical Analysis}.

\paragraph{Neural Network for the Data Driven Approach}

The baseline approach is based on training a Neural Network to extend a partial assignment (also called a \emph{partial solution}) by making one additional assignment, so as to preserve feasibility. Formally, the network is a function:
\begin{equation}
   f: \{0, 1\}^m \rightarrow [0,1]^m
\end{equation}
whose input and output are $m$ dimensional vectors. Each element in the vectors is associated to a variable-value pair $\langle z_j, v_j \rangle$, where $z_j$ is the associated variable and $v_j$ is the associated value. The network input represents partial assignments, assuming that $x_j = 1$ iff $z_j = v_j$. Each component $f_j(x)$ of the output is proportional to the probability that pair $\langle z_j, v_j \rangle$ is chosen for the next assignment. This is achieved in practice by using an output layer with $m$ neurons with a sigmoid activation function.


\paragraph{Dataset Generation Process}

The input of each training example corresponds to a partial solution $x$, and the output to a single variable value assignment (represented as a vector $y$ using a one-hot encoding). The training set is constructed by repeatedly calling the randomized deconstruction procedure \Cref{alg:deconstruct} on an initial set of full solutions (referred to as \emph{solution pool}). Each call generates a number of examples that are used to populate a dataset. At the end of the process we discard multiple copies of identical examples. Two examples may have the same input, but different output, since a single partial assignment may have multiple viable completions.


Unlike \cite{galassi2018-01}, here we sometimes perform \emph{multiple calls to \Cref{alg:deconstruct} for the same starting solution}. This simple approach enables to investigate independently the effect of the training set size (which depends on the number of examples) and of the amount of actual empirical knowledge (which depends on the size of the solution pool). The method also enables to generate large training sets starting from a very small number of historical solutions.

\begin{algorithm}[tb]
   \small
   \caption{\sc deconstruct($x$)}
   \begin{algorithmic}
      \STATE $D = \emptyset$
      \WHILE{$\|x\|_1 > 0$}
      \STATE Let $y = \mathbf{0}$ \quad \emph{\# zero vector}
      \STATE Select a random index $i$ s.t. $x_i = 1$
      \STATE Set $x_i = 0$, set $y_i = 1$
      \STATE Add the pair $\langle x, y \rangle$ to $D$
      \ENDWHILE
      \RETURN $D$
   \end{algorithmic}
   \label{alg:deconstruct}
\end{algorithm}

\paragraph{Training and Knowledge Injection}

The basic training for the NN is the same as for neural classifiers. Since the network output can be assimilated to a class, we process the network output through a softmax operator, and then we use as a loss function the categorical crossentropy $H$. Additionally, we inject domain knowledge at training time via an approach that combines ideas of Semantic Based Regularization (SBR) and Constraint Programming (CP, \cite{rossi2006handbook}).

In CP, constraints are associated to algorithms called \emph{propagators} that can identify provably infeasible values in the domain of the variables. Propagators are generally incomplete, i.e. there is no guarantee they will find \emph{all} infeasible values. Given a constraint (or a collection of constraints) $C$, here we will treat its propagator as a multivariate function such that $C_j(x) = 1$ iff assignment $z_j = v_j$ has not been marked as infeasible by the propagator, while $C_j(x) = 0$ otherwise. We then augment the loss function with a SBR inspired term that discourages provably infeasible pairs, and encourages the remaining ones. Given an example $\langle x, y \rangle$, we have:
\begin{equation}
   L_{\mathit{sbr}}(x) = \sum_{j = 0}^{m-1} (C_j(x) - f_j(x))^2
   \label{eqn:sbr_loss}
\end{equation}
i.e. increasing the output of a neuron corresponding to a provably infeasible pair incurs a penalty that grows with the square of $f_j(x)$; increasing the output for the remaining pairs incurs a penalty that grows with the square of $1-f_j(x)$. Our full loss is hence given by:
\begin{equation}
   H\left(\frac{1}{Z}f(x), y\right) + \lambda L_{\mathit{sbr}}(x)
\end{equation}
where $Z$ is the partition function and the scalar $\lambda$ controls the balance between the crossentropy term $H$ and the SBR term. \emph{The method can be applied for all known propagators with discrete variables}. With some adaptations, it can be made to work for important classes of numerical propagators (e.g. those that enforce Bound Consistency \cite{rossi2006handbook}).

Since propagators are incomplete and we are rewarding assignments not marked as infeasible, there is a chance that our SBR term injects incorrect information in the model. This reward mechanism is however crucial to ensure a non-negligible gradient at training time: the incorrect information is balanced by the presence of the crossentropy term, which encourages assignments that are guaranteed feasible (since they originate from full problem solutions).


\paragraph{Evaluation and Knowledge Injection}

We evaluate the approach via a specialized procedure, relying on a randomized solution function for PLS instances. This has signature \textsc{solve}($x$, $C$, $h$), where $x$ is the starting partial assignment, $C$ is the considered (sub)set of problem constraints, and $h$ is a probability estimator for variable-value pairs (e.g. our trained NN). The function is implemented via the Google or-tools constraint solver, and is based on Depth First Search: at each search node, the solver applies constraint propagation to prune some infeasible values, it chooses for branching the first unassigned variable in a static order, then assigns a value chosen at random with probabilities proportional to $h(x^\prime)$, where $x^\prime$ is the current state of assignments. The \textsc{solve} function returns either a solution, or $\bot$ in case of infeasibility.

Our evaluation method tests the ability of the NN to identify individual assignments that are globally feasible, i.e. that can be extended into full solutions. This is done via \Cref{alg:feastest}, which 1) starts from a given partial solution; 2) relies on a constraint propagator $C$ to discard some of the provably infeasible assignments; 3) uses the NN to make a (deterministic) single assignment; 4) attempts to complete it into a full solution (taking into account all problem constraints, i.e. $C_{\mathit{pls}}$). Replacing the NN with a uniform probability estimator allows to obtain a baseline. We repeat the process on all partial solutions from a test set, and collect statistics. This approach is identical to one of those in \cite{galassi2018-01}, with one major difference, i.e. the use of a constraint propagator for ``correcting'' the output of the probability estimator. This enables injection of (incomplete) knowledge at solution construction time, while the original behavior can be recovered by using an empty set of propagators.

\begin{algorithm}
   \small
   \caption{\textsc{feastest}($X$, $C$, $h$)}
   \begin{algorithmic}
      \STATE $J^* = \arg\max \{h_j(x) \mid C_j(x) = 1\}$ \COMMENT{Most likely assignments}
      \STATE Pick $j^*$ uniformly at random from $J^*$
      \STATE Set $x_{j^*} = 1$
      \IF{\textsc{solve}($x$, $C_{\mathit{pls}}$, $h_{\mathit{rnd}}$) $\neq \bot$}
      \RETURN 1 \COMMENT{Globally feasible}
      \ELSE
      \RETURN 0 \COMMENT{Globally infeasible}
      \ENDIF
   \end{algorithmic}
   \label{alg:feastest}
\end{algorithm}


Unlike in typical Machine Learning evaluations, we do not measure the (extremely low) network accuracy: in fact, the accuracy metric in our case is tied to the network ability to replicate the same sequence of assignments observed at training time, which has almost no practical value.

\section{Empirical Analysis}%
\label{sec:Empirical Analysis}

In this section we discuss our experimental analysis, which is designed around some key questions of both scientific and practical import. We focus on the following aspects:
\begin{description}
   \item[Q1:] Does injecting knowledge at training time improve the network ability to identify feasible assignments?
   \item[Q2:] What is the effect of injecting domain knowledge after, rather than during, training?
   \item[Q3:] What is the effect of adjusting the amount of available empirical knowledge?
\end{description}
Taking advantage of our controlled use case, we present a series of experiments that investigate such research directions. Details about the rationale and the setup of each experiment are reported in dedicated sections, but some common configuration can be immediately described.

We focus on $10\times 10$ PLS instances, resulting in input and output vectors with $1,000$ elements. We use a feed-forward, fully-connected Neural Network with three hidden layers, each with 512 units having ReLU activation function. This setup is considerably simpler than the one used in the approach we chose as our baseline, but manages to reach very similar results. We employ the Adam optimizer from Keras-TensorFlow2.0, with default parameters. The number of training epochs and batch size depends on the experiment. 

\subsection{Domain Knowledge at Training Time}%
\label{sub:Domain Knowledge at Training Time}

We start with an experiment designed to address Question 1, i.e. whether injecting domain knowledge \emph{at training time} may help the NN in the identification of feasible assignments. This is practically motivated by situations in which a domain expert has only partial information about the problem structure, but a pool of historical solutions is available.

For this experiment, the training set for the network is generated using the deconstruction approach from \Cref{sec:Methods}, starting from a set of 10,000 PLS solutions. Each solution is then deconstructed exactly once, yielding a training set of $\sim$ 730,000 examples, 25\% of which are then removed to form a separate test set. We use mini-batches of 50,000 examples and we stop training after 1000 epochs.

We compare four approaches: a random approach (referred to as \textsc{rnd}), which treats all possible variable-value pairs as equally likely; a model-agnostic neural network (referred to as \textsc{agn}); a network trained with knowledge about row constraints (referred to as \textsc{rows}); a network trained with knowledge about row and column constraints (referred to as \textsc{full}).

The \textsc{rnd} approach lacks even the basic knowledge that a variable cannot be assigned twice, since this is not enforced by our input/output encoding. The same holds for \textsc{agn}, which can however infer such constraint from data. Conversely, in \textsc{rows} we use our SBR-inspired method (and a Forward Checking propagator) to inject knowledge that both assigning a variable twice and assigning a value twice on the same row is forbidden. For the \textsc{full} approach we do the same, applying the Forward Checking propagator also to column constraints (i.e. no value can appear twice on the same column). Due to the use of an incomplete propagator, both \textsc{row} and \textsc{full} make use of incomplete knowledge. We have empirical determined that $\lambda = 1$ for \textsc{full} and $\lambda = 0.01$ for \textsc{row} works reasonably well.

\begin{figure}[tb]
\begin{center}
\includegraphics[width=8cm]{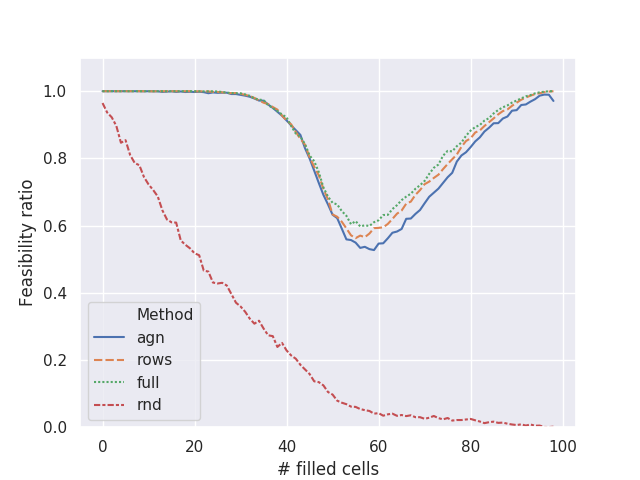}
\caption[Knowledge Injection at Training Time]{Knowledge Injection at Training Time}
\label{fig:prima}
\end{center}
\end{figure}

We evaluate the resulting approaches via the \textsc{feastest} procedure, using the (separated) test set as $X$, the trained networks (or the uniformly random estimator) as $h$, and an empty set of constraints (i.e. no propagation at test time). We then produce ``feasibility plots'' that report on the x-axis the number of assigned variables (filled cells) in the considered partial solutions and on the y-axis the ratio of suggested assignments that are globally feasible.

The results of the evaluation are shown in \Cref{fig:prima}. Without propagating any constraint at evaluation time, a purely random choice is extremely unlikely to result in globally feasible assignments: this is expected and only serves as a pessimistic baseline. The \textsc{agn} approach, relying exclusively on empirical knowledge, behaves considerably better, with high feasibility ratios for almost empty and almost full squares, and a noticeable drop when $\sim$60\% of the square is filled. The trend is a common feature for many of the approaches, and may be connected a known phase transition in the complexity of this combinatorial problem. \emph{Injecting domain knowledge at training time improves the feasibility ratio by a noticeable margin}: a half of the improvement is observed when moving from \textsc{agn} to \textsc{row}, suggesting that even partial knowledge about the problem structure could prove very useful.

\subsection{Domain Knowledge at Evaluation Time}%
\label{sub:Domain Knowledge at Evaluation Time}

Our second experiment addresses Question 2, regarding the effect of knowledge injection at test time. This topic relates to scenarios where the expert-supplied information can be incorporated in a solution method (e.g. as a constraint in a declarative model). While not always viable, this situation is frequent enough to deserve a dedicated analysis.

For this experiment, we rely on the same training and test set as in \Cref{sub:Domain Knowledge at Training Time}, and compare the same approaches. As a main difference, we take into account some or all the problem constraints at evaluation time, by passing a non-empty set $C$ of propagators to the \textsc{feastest} procedure. Also at test time the constraints are taken into account by means of an incomplete propagator (Forward Checking), and hence all approaches rely on incomplete knowledge.

\begin{figure}[tb]
\begin{center}
\includegraphics[width=8cm]{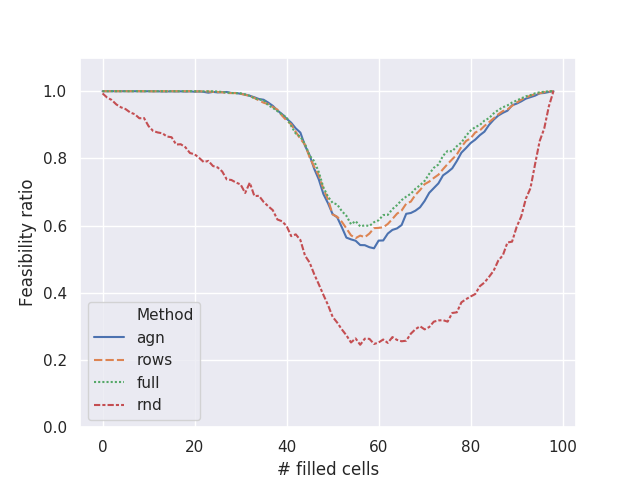}
\caption[Knowledge Injection at Evaluation Time (\textsc{rows})]{Knowledge Injection at Evaluation Time (\textsc{rows})}
\label{fig:seconda}
\end{center}
\end{figure}

\begin{figure}[tb]
\begin{center}
\includegraphics[width=8cm]{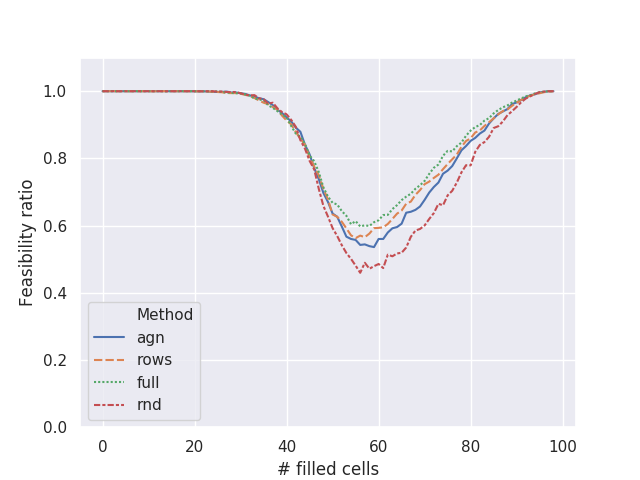}
\caption[Knowledge Injection at Evaluation Time (\textsc{full})]{Knowledge Injection at Evaluation Time (\textsc{full})}
\label{fig:terza}
\end{center}
\end{figure}

The results of this evaluation are presented in \Cref{fig:seconda} (for row constraints propagation at test time) and \Cref{fig:terza} (for all problem constraints). The \textsc{rnd} results in this case are representative of the behavior (at each search node) of a Constraint Programming solver having access to either only row constraints or the full problem definition: since CP is known to work very well on the PLS, it is therefore expected that the performance of the random approach is significantly boosted by knowledge injection at test time.

All approaches relying either on purely (\textsc{agn}) or partially (\textsc{rows} and \textsc{full}) on empirical knowledge gain almost no benefit from injecting constraints at evaluation time, though they still perform noticeably better than \textsc{rnd}. On one hand, these diminishing returns should be taken into account when taking into account constraints during solution generation is viable. On the other hand, the fact that all knowledge-driven approaches are not helped by constraint propagation suggests that \emph{their advantage comes from information about global feasibility, which they can access from the empirical data}.

\subsection{Training Set Size and Empirical Information}%
\label{sub:Training Set Size and Empirical Information}

Next, we proceed to tackle Question 3, by acting on the training set generation process. In classical Machine Learning approaches, the amount of available information is usually measured via the training set size: this is a reasonable approach, since the number of training examples has a strong impact on the ability of a ML method to learn and generalize.

\begin{figure}[tb]
\begin{center}
\includegraphics[width=8cm]{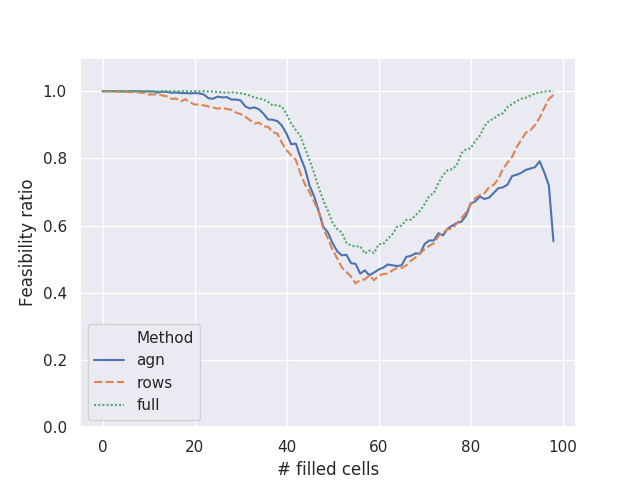}
\caption[Effect of Training Set Size (300k examples)]{Effect of Training Set Size (300k examples)}
\label{fig:quarta}
\end{center}
\end{figure}

\begin{figure}[tb]
\begin{center}
\includegraphics[width=8cm]{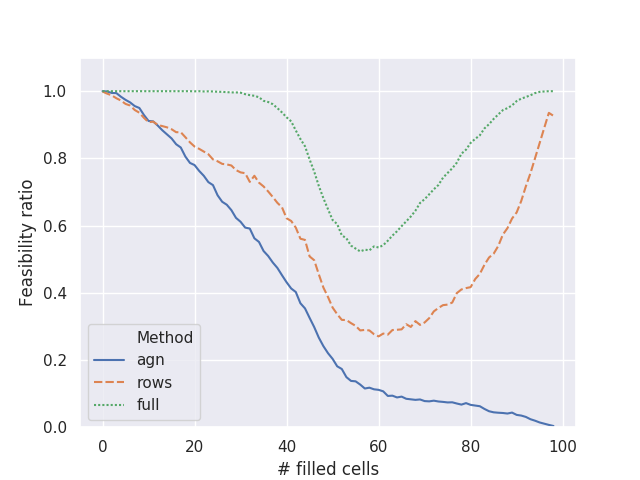}
\caption[Effect of Training Set Size (50k examples)]{Effect of Training Set Size (50k examples)}
\label{fig:quinta}
\end{center}
\end{figure}

We performed experiments to probe the effect of the training set size on the performance of the data-driven approaches. \Cref{fig:quarta} and \Cref{fig:quinta} report results for training sets with respectively 300,000 and 50,000 examples. \emph{Knowledge injection at training time has in this case a dramatic effect}: while the \textsc{agn} approach is very sensitive to the available number of examples, the \textsc{full} one has only a minor drop in performance when moving from $\sim$730,000 examples to 50,000 examples. This confirms previous experiences with techniques such as Semantic Based Regularization, although the effect in our case is much more pronounced: the gap is likely due to the fact that, while our cross-entropy term in the loss function provides information about a single (globally) feasible assignment, the SBR terms provides information about a large number of (locally) feasible assignments.

In our setup, we have also the possibility to apply the deconstruction process multiple times, so that the number of different examples that can be obtained from a single solution grows with the number of possible permutations of the variable indices (i.e. $O(n^2!)$ for the PLS). Such observation opens up the possibility to generate large training sets from a very small number of starting solutions: this is scientifically interesting, since the ``actual'' empirical information depends on how many solutions are available; this is also very useful in practice, since in many practical applications only a relatively small number of historical solutions exists.

\begin{figure}[tb]
\begin{center}
\includegraphics[width=8cm]{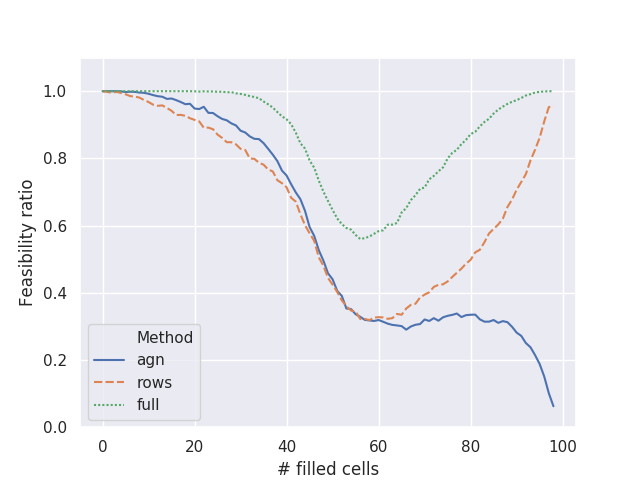}
\caption[Effect of Solution Pool Size (1k solutions)]{Effect of Solution Pool Size (1k solutions)}
\label{fig:sesta}
\end{center}
\end{figure}

\begin{figure}[tb]
\begin{center}
\includegraphics[width=8cm]{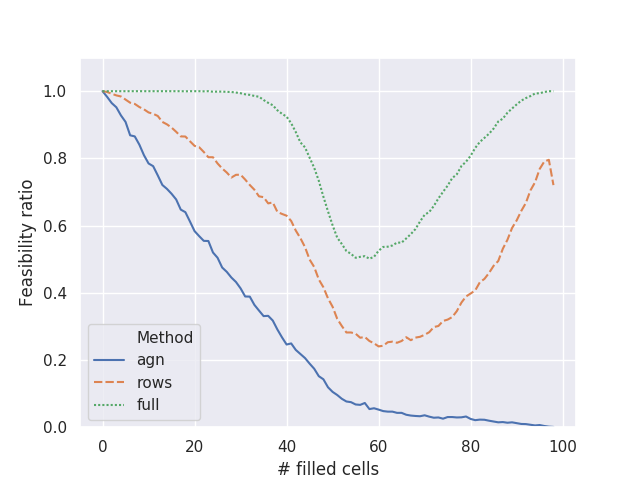}
\caption[Effect of Solution Pool Size (100 solutions)]{Effect of Solution Pool Size (100 solutions)}
\label{fig:settima}
\end{center}
\end{figure}

The results of this evaluation are shown in \Cref{fig:sesta} and \Cref{fig:settima} for solution pools having respectively 1,000 and 100 elements. In both cases the size of the generated training set is comparable to the original, i.e. around 700,000 examples: despite this fact, there is a very significant gap in performance between the \textsc{agn} approach and \textsc{full}. This is likely due once again to the richer information made accessible via the combined use of propagators and our SBR-inspired loss.

From a practical point of view it seems that, \emph{as long as enough problem knowledge is available, it is possible to train data-driven methods with very high feasibility ratio, starting from very small pools of historical solutions}. It may be argued that if extensive problem knowledge is available, one may use a more traditional solution approach (e.g. Constraint Programming or Mathematical Programming): even in such a case, however, a (partially) data-driven approach should have a higher chance to preserve implicit properties (e.g. user preferences) that make the historical solutions desirable.

\section{Conclusion}%
\label{sec:Conclusion}

We considered injecting domain knowledge in Deep Neural Networks to bridge the gap between expert-designed models and data-driven approaches for constrained problems. We chose the PLS as a case study, and extended an existing NN approach to enable knowledge injection. We performed controlled experiments to investigate three main scientific questions, drawing the following conclusions:
\begin{description}
   \item[Q1:] Injecting domain knowledge at training time improves the ability of the NN approach to identify feasible assignments. Data driven methods behave significantly better than a naive random baseline.
   \item[Q2:] Using constraint propagation to filter out some infeasible assignments at test time improves dramatically the behavior of random selection; data-driven methods receive almost no benefit, but they still perform best. This suggests that data-driven methods can infer information about global feasibility from empirical data.
   \item[Q3:] A pure data-driven approach is very sensitive to the available empirical information. Injecting knowledge at training time improves robustness: if both row and column constraints are considered, a limited performance drop is observed with as few as 100 historical solutions.
\end{description}
Our analysis required the development of a general, SBR-inspired, method to turn any constraint propagator into a source of training-time information. Due to this result, together with our conclusions, this paper makes a significant step toward the practical applicability of data-driven approaches for constraint problems.

Many open questions remain: an experimentation with different problem types and scales is needed to make sure that our results hold in general. Embedding the NN in an actual search process (with or without propagation) will provide more insight into the global behavior of the data-driven methods; finally, when applying propagation at training time is viable, it is desirable to adjust training so that it complements, rather than replicate, the effect of propagation.


\newpage

\bibliographystyle{named}
\bibliography{ijcai20}

\end{document}